\documentclass[sigconf]{acmart}
\AtBeginDocument{%
  \providecommand\BibTeX{{%
    \normalfont B\kern-0.5em{\scshape i\kern-0.25em b}\kern-0.8em\TeX}}}

\usepackage{algorithm}
\usepackage{algorithmic}

\usepackage{amsfonts}
\usepackage{amsmath}
\usepackage{amssymb}
\usepackage{multirow}
\usepackage{float}
\newcommand\numberthis{\addtocounter{equation}{1}\tag{\theequation}}
\usepackage{graphicx}

\newtheorem{Remark}{Remark}

\copyrightyear{2022}
\acmYear{2022}
\setcopyright{othergov}\acmConference[KDD '22]{Proceedings of the 28th ACM SIGKDD Conference on Knowledge Discovery and Data Mining}{August 14--18,2022}{Washington, DC, USA}
\acmBooktitle{Proceedings of the 28th ACM SIGKDD Conference on Knowledge Discovery and Data Mining (KDD '22), August 14--18, 2022, Washington, DC, USA}
\acmPrice{15.00}
\acmDOI{10.1145/3534678.3539254}
\acmISBN{978-1-4503-9385-0/22/08}

\begin{document}
\title{Connecting Low-Loss Subspace for Personalized Federated Learning}

\author{Seok-Ju Hahn}
\email{seokjuhahn@unist.ac.kr}
\affiliation{%
  \institution{Ulsan National Institute of Science and Technology \& Kakao Enterprise}
  \country{South Korea}
}
\author{Minwoo Jeong}
\email{lloyd.ai@kakaoenterprise.com}
\affiliation{%
  \institution{Kakao Enterprise}
  \country{South Korea}
}
\author{Junghye Lee}
\email{Junghyelee@unist.ac.kr}
\affiliation{%
  \institution{Ulsan National Institute of Science and Technology}
  \country{South Korea}
}

\begin{abstract}
   Due to the curse of statistical heterogeneity across clients, adopting a personalized federated learning method has become an essential choice for the successful deployment of federated learning-based services. Among diverse branches of personalization techniques, a model mixture-based personalization method is preferred as each client has their own personalized model as a result of federated learning. It usually requires a local model and a federated model, but this approach is either limited to partial parameter exchange or requires additional local updates, each of which is helpless to novel clients and burdensome to the client's computational capacity. As the existence of a connected subspace containing diverse low-loss solutions between two or more independent deep networks has been discovered, we combined this interesting property with the model mixture-based personalized federated learning method for improved performance of personalization. We proposed SuPerFed, a personalized federated learning method that induces an explicit connection between the optima of the local and the federated model in weight space for boosting each other. Through extensive experiments on several benchmark datasets, we demonstrated that our method achieves consistent gains in both personalization performance and robustness to problematic scenarios possible in realistic services.
\end{abstract}

\ccsdesc[500]{Computing methodologies~Distributed algorithms}
\ccsdesc[300]{Computing methodologies~Neural networks}

\keywords{federated learning, non-IID data, personalization, personalized federated learning, label noise, mode connectivity}
\maketitle
\section{Introduction}
Individuals and institutions are now data producers as well as data keepers, thanks to advanced communication and computation technologies. Therefore, training a machine learning model in a data-centralized setting is sometimes not viable due to many realistic limitations, such as the existence of massive clients generating their own data in real-time or the data privacy issue restricting the collection of data. Federated learning (FL)~\cite{mc+17} is a solution to solve this problematic situation as it enables parallel training of a machine learning model across clients or institutions without sharing their private data, usually under the orchestration of the server (e.g., service providers). In the most common FL setting, such as \texttt{FedAvg}~\cite{mc+17}, when the server prepares and broadcasts a model appropriate for a target task, each participant trains the model with its own data and transmits the resulting model parameters to the server. Then, the server receives the locally updated model parameters to aggregate (i.e., weighted averaging) into a new global model. The new global model is broadcast again to some fraction of participating clients. This collaborative learning process is repeated until convergence. 
\begin{figure}
\centering
\includegraphics[width=0.8\linewidth]{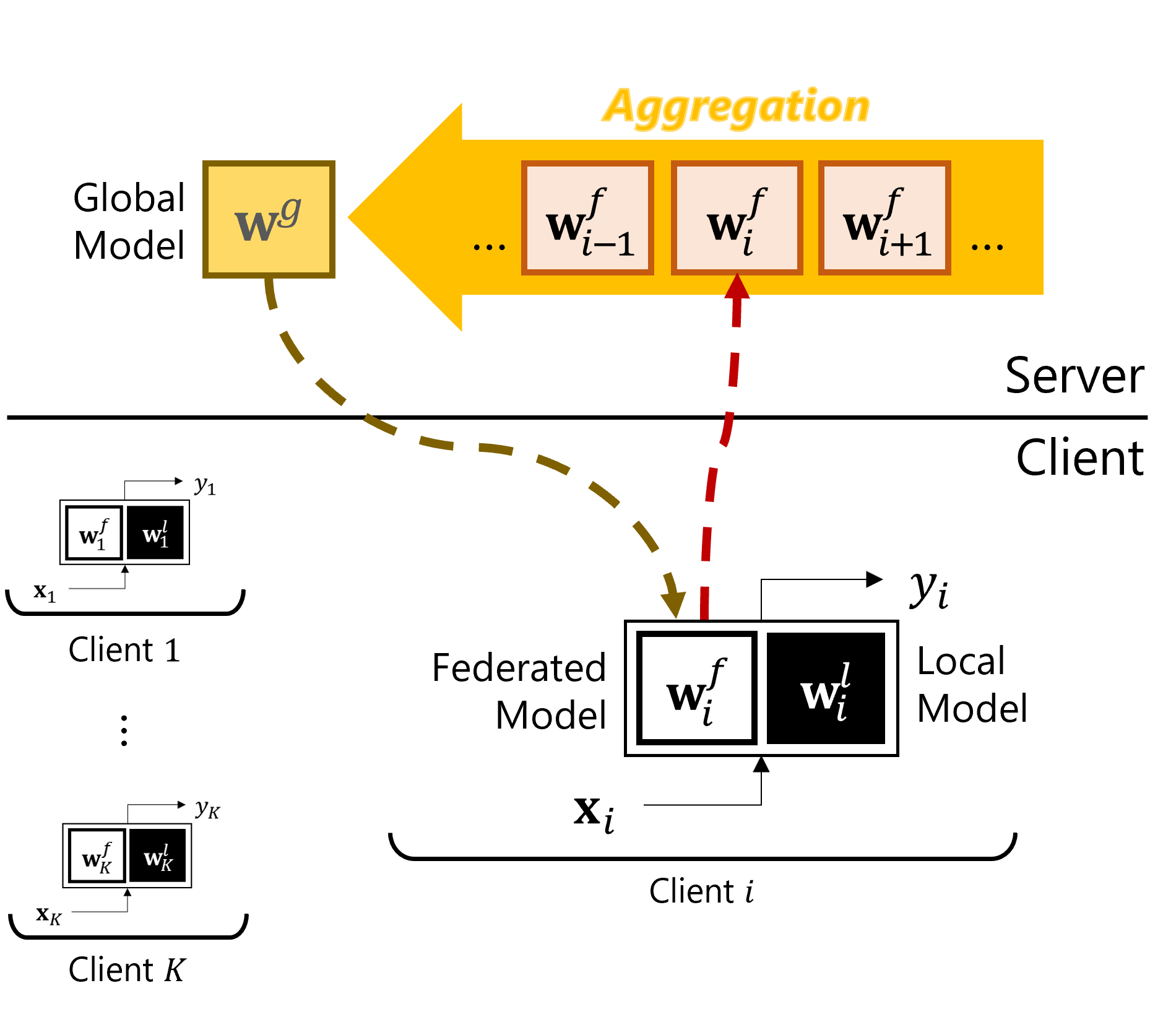}
\caption{Overview of the model mixture-based personalized federated learning.}
\label{fig1}
\end{figure}
However, obtaining a single global model through FL is not enough to provide satisfiable experiences to \textit{all} clients, due to the curse of data heterogeneity. Since there exists an inherent difference among clients, the data residing in each client is not statistically identical. That is, a different client has their own different data distribution, or the data is not independent and identically distributed (non-IID). Thus, applying a simple FL method like \texttt{FedAvg} cannot avoid a pitfall of the high generalization error or even the divergence of model training~\cite{ka+19}. That is, it is sometimes possible that the performance of the global model is lower than the locally trained model without collaboration. When this occurs, clients barely have the motivation to participate in the collaborative learning process to train a single global model.

Hence, it is natural to rethink the scheme of FL. As a result, personalized FL (PFL) has becomes an essential component of FL, which has yielded many related studies. Many approaches are proposed for PFL, based on multi-task learning \cite{mocha, mar+21, fedu}, regularization technique \cite{l2sgd, pfedme, ditto}; meta-learning  \cite{Fallah, jiang+19}, clustering \cite{ghosh+20, clustered, mansour+20}, and a model mixture method \cite{fedrep, FedPer, lgfedavg, ditto, apfl, mansour+20}. We focus on the model mixture-based PFL method in this paper as it shows promising performance as well as lets each client have their own model, which is a favorable trait in terms of personalization as each participating client eventually has their own model. This method assumes that each client has two distinct parts: a local (sub-)model and a federated (sub-)model\footnote{We intentionally coined this term to avoid confusion with the global model at the server.} (i.e., federated (sub-)model is a global model transferred to a client). In the model mixture-based PFL method, it is expected that the local model captures the information of heterogeneous client data distribution by staying only at the client-side, while the federated model focuses on learning common information across clients by being communicated with the server. Each federated model built at the client is sent to the server, where it is aggregated into a single global model (Figure~\ref{fig1}). At the end of each PFL iteration, a participating client ends up having a personalized part (i.e., a local model), which potentially relieves the risks from the non-IIDness as well as benefits from the collaboration of other clients.

The model mixture-based PFL method has two sub-branches. One is sharing only a partial set of parameters of one single model as a federated model \cite{fedrep, FedPer, lgfedavg} (e.g., only exchanging weights of a penultimate layer), the other is having two identically structured models as a local and a federated model \cite{ditto, apfl, l2sgd, pfedme, mansour+20}. However, each of them has some limitations. In the former case, new clients cannot immediately exploit the model trained by FL, as the server only has a partial set of model parameters. In the latter case, all of them require separate (or sequential) updates for the local and the federated model each, which is possibly burdensome to some clients having low computational capacity.

Here comes our main motivation: the development of a model mixture-based PFL method that can jointly train a whole model communicated with the server and another model for personalization. We attempted to achieve this goal through the lens of \textit{connectivity}. As many studies on the deep network loss landscape have progressed, one intriguing phenomenon is being actively discussed in the field: the \textit{connectivity} \cite{garipov+18, fort2019deep, fort2019large, draxler18, modeconnect, nnsubspaces} between deep networks. Though deep networks are known to have many local minima, it has been revealed that each of these local minima has a similar performance to each other \cite{choro}. Additionally, it has recently been discovered that two different local minima obtained by two independent deep networks can be connected through the linear path \cite{frankle20} or the non-linear path \cite{garipov+18, draxler18} in the weight space, where \textit{all} weights on the path have a low loss. Note that the two endpoints of the path are two different local minima reached from the two different deep networks. These findings have been extended to a multi-dimensional subspace \cite{nnsubspaces, benton21} with only a few gradient steps on two or more independently initialized deep networks. The resulting subspace, including the one-dimensional connected path, contains functionally diverse models having high accuracy. It can also be viewed as an ensemble of deep networks in the weight space, thereby sharing similar properties such as good calibration performance and robustness to the label noise. By introducing the \textit{connectivity} to a model mixture-based PFL method, such good properties can also be absorbed advantageously.

\medskip
\noindent {\bf{Our contributions.}}
Inspired by this observation, we propose \texttt{SuPerFed}, a connected low-loss \textbf{su}bspace construction method for the \textbf{per}sonalized \textbf{fed}erated learning, adapting the concept of \textit{connectivity} to the model mixture-based PFL method. This method aims to find a low-loss subspace between a single global model and many different local models at clients in a mutually beneficial way. Accordingly, we aim to achieve better personalization performance while overcoming the aforementioned limitations of model mixture-based PFL methods. Adopting the \textit{connectivity} to FL is non-trivial as it obviously differs from the setting where the \textit{connectivity} is first considered: each deep network observes different data distributions and only the global model can be connected to other local models on clients.
The main contributions of the work are as follows:
\begin{description}
\item[$\bullet$] We propose \texttt{SuPerFed}, a model mixture-based PFL method that aims to achieve good personalization performance of each local model on clients by harmoniously connecting them with the global model as a pivot, under diverse non-IID scenarios.
\item[$\bullet$] Our proposed method extends the limitations of existing model mixture-based PFL methods by jointly updating both federated and local models in clients and exchanging the entire federated model with the server, allowing novel clients to benefit from it.
\item[$\bullet$] Personalized models trained by adopting our method are well-calibrated and also robust to potential label noise that is a common problem in realistic FL-driven services.
\end{description}
\begin{figure*}[!htb]
\centering
\includegraphics[width=1.0\linewidth]{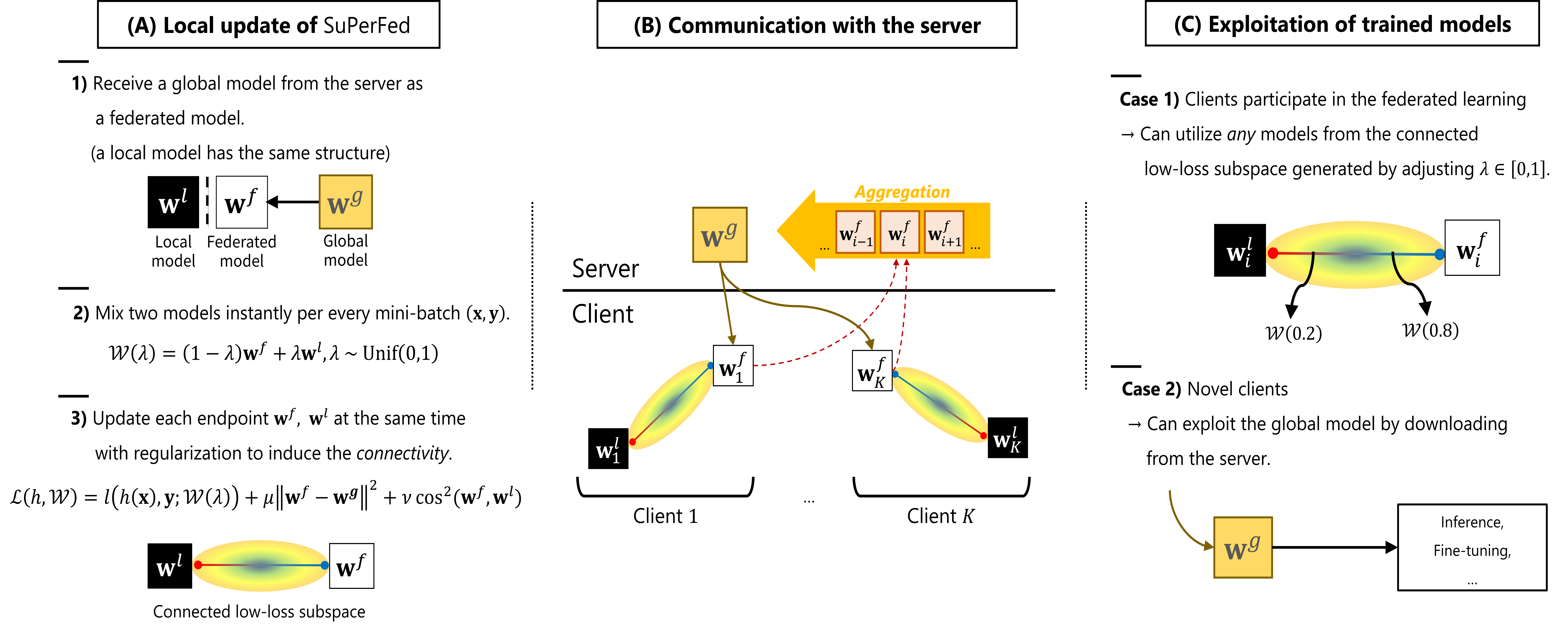}
\caption{An illustration of the proposed method \texttt{SuPerFed}. (A) Local update of \texttt{SuPerFed}: at every federated learning round, a selected client receives a global model from the server and sets it to be a federated model. After being mixed by a randomly generated $\lambda$, two models are jointly updated with regularization. (B) Communication with the server: only the updated federated model is uploaded to the server (dotted arrow in crimson) to be aggregated (e.g., weighted averaging) as a new global model, and it is broadcast to clients in the next round (arrow in gray). (C) Exploitation of trained models: (Case 1) FL clients can sample and use \textit{any} model on the connected subspace (e.g., $\mathcal{W}(0.2), \mathcal{W}(0.8)$) because it only contains low-loss solutions. (Case 2) Novel clients can download and use the server's trained global model $\mathbf{w}_g$.}\label{fig2}
\end{figure*}

\section{Related Works}
\subsection{FL with Non-IID Data} After \cite{mc+17} proposed the basic FL algorithm (\texttt{FedAvg}), handling non-IID data across clients is one of the major points to be resolved in the FL field. While some methods are proposed such as sharing a subset of client's local data at the server\cite{zhao+18}, accumulating previous model updates at the server \cite{fedavgm}. These are either unrealistic assumptions for FL or not enough to handle a realistic level of statistical heterogeneity. Other branches to aid the stable convergence of a single global model include modifying a model aggregation method at the server \cite{pfnm, fedma, pillu+19, fedbn} and adding a regularization to the optimization \cite{fedprox, ka+19, feddyn}. However, a single global model may still not be sufficient to provide a satisfactory experience to clients using FL-driven services in practice.
\subsection{PFL Methods} As an extension of the above, PFL methods shed light on the new perspective of FL. PFL aims to learn a client-specific personalized model, and many methodologies for PFL have been proposed.
Multi-task learning-based PFL \cite{mocha, fedu, mar+21} treats each client as a different task and learns a personalized model for each client. Local fine-tuning-based PFL methods \cite{jiang+19, Fallah} adopt a meta-learning approach for a global model to be adapted promptly to a personalized model for each client, and clustering-based PFL methods \cite{ghosh+20, clustered, mansour+20} mainly assume that similar clients may reach a similar optimal global model. Model mixture-based PFL methods \cite{apfl, FedPer, lgfedavg, mansour+20, pfedme, fedrep, ditto, l2sgd} divide a model into two parts: one (i.e., a local model) for capturing local knowledge, and the other (i.e., a federated model) for learning common knowledge across clients. In these methods, only the federated model is shared with the server while the local model resides locally on each client. \cite{FedPer} keeps weights of the last layer as a local model (i.e., the personalization layers), and \cite{fedrep} is similar to this except it requires a separate update on a local model before the update of a federated model. In contrast, \cite{lgfedavg} retains lower layer weights in clients and only exchanges higher layer weights with the server. Due to the partial exchange, new clients in the scheme of \cite{FedPer, fedrep, lgfedavg} should train their models from scratch or need at least some steps of fine-tuning. In \cite{mansour+20, apfl, l2sgd, ditto, pfedme}, each client holds at least two separate models with the same structure:   one for a local model and the other for a federated model. In \cite{mansour+20, apfl,l2sgd}, they are explicitly interpolated in the form of a convex combination after the independent update of two models. In \cite{ditto, pfedme}, the federated model affects the update of the local model in the form of proximity regularization.
\subsection{Connectivity of Deep Networks} The existence of either linear path \cite{frankle20} or non-linear path \cite{garipov+18, draxler18} between two minima derived by two different deep networks has been recently discovered through extensive studies on the loss landscape of deep networks \cite{fort2019deep, fort2019large, modeconnect, garipov+18, draxler18}. That is, there exists a low-loss subspace (e.g., line, simplex) connecting two or more deep networks independently trained on the same data. Though there exist some studies on constructing such a low loss subspace \cite{blundell+15, snapshot, swa, swa-gaussian, izmailov2020subspace, deepensembles}, they require multiple updates of deep networks. After it is observed that independently trained deep networks have a low cosine similarity with each other in the weight space \cite{fort2019deep}, a recent study by \cite{nnsubspaces} proposes a straightforward and efficient method for explicitly inducing linear connectivity in a single training run. Our method adopts this technique and adapts it to construct a low-loss subspace between each local model and a federated model (which will later be aggregated into a global model at the server) suited for an effective PFL in heterogeneously distributed data.

\section{Proposed Method}
\begin{algorithm}[!tb]
\caption{LocalUpdate}
\label{alg:LocalUpdate}
\begin{algorithmic}
\STATE{\textbf{Inputs}: global model from the server, $\mathbf{w}^g$, batch size $\mathrm{B}$, number of local epochs $\mathrm{E}$, current round $r$, start round of personalization $\mathrm{L}$, learning rate $\eta$, regularization constants $\mu, \nu$, client dataset $\mathcal{D}$.}
\STATE{\textbf{Start:}} set the federated model as: $\mathbf{w}^f\leftarrow\mathbf{w}^g$.
\IF{the local model $\mathbf{w}^l$ does not exist}
\STATE Set the local model as: $\mathbf{w}^l\leftarrow\mathbf{w}^g$.
\ENDIF
\FOR{$e=0,...,\mathrm{E}-1$}
\STATE $\mathcal{B}_e\leftarrow$Split the client dataset $\mathcal{D}$ into batches of size $\mathrm{B}$.
\FOR{a local batch $(\mathbf{x},\mathbf{y})\in\mathcal{B}_e$}
\IF{$r<\mathrm{L}$}
\STATE Set $\lambda=0$
\ELSE
\STATE Sample $\lambda \sim \text{Unif}(0,1)$.
\ENDIF
\STATE Mix models $\mathcal{W}(\lambda)=(1-\lambda)\mathbf{w}^f+\lambda\mathbf{w}^l$.
\STATE Set the local objective $\mathcal{L}$ using (3) and (4).
\STATE Minimize $\mathcal{L}$ in terms of $\mathbf{w}^f$ (5) and $\mathbf{w}^l$ (6) each through $\mathcal{W}(\lambda)$ using SGD with the learning rate $\eta$.
\ENDFOR
\ENDFOR
\STATE{\textbf{Return:}} updated federated model $\mathbf{w}^f$.
\end{algorithmic}
\end{algorithm}
\begin{algorithm}[!tb]
\caption{\texttt{SuPerFed}}
\label{alg:SuPerFed}
\begin{algorithmic}
\STATE{\textbf{Inputs}}: batch size $\mathrm{B}$, number of local epochs $\mathrm{E}$, total communication rounds $\mathrm{R}$, start round of personalization $\mathrm{L}$, learning rate $\eta$, regularization constants $\mu, \nu$, number of clients $\mathrm{K}$, fraction of clients to be sampled $\mathrm{C}$, clients $c_i$ having own dataset $\mathcal{D}_i=\{(\mathbf{x}_i^j,\mathbf{y}_i^j)\}_{j=1}^{n_i}, i\in[\mathrm{K}]$ \\ 
\STATE{\textbf{Start}:} Server initializes a global model $\mathbf{w}^{g, 0}$.
\FOR{$r=0,..,\mathrm{R}-1$}
\STATE Server randomly selects $\max(\mathrm{C}\cdot\mathrm{K}, 1)$ clients as $S_r$.
\STATE Server broadcasts the current global model $\mathbf{w}^{g,r}$ to $S_r$.
\FOR{each client $c_i\in S_r$ \textbf{in parallel}} 
    \STATE $\mathbf{w}^{f, r}_i\leftarrow$ \text{LocalUpdate($\mathbf{w}^{g, r}, \mathrm{B}, \mathrm{E}, r, \mathrm{L}, \eta, \mu, \nu$)}
\ENDFOR
\STATE Update a global model: \\
$\mathbf{w}^{g, r+1}\leftarrow\frac{1}{\sum_{i\in S_r}{n_i}}\sum_{i\in S_r}{n_i}\mathbf{w}_i^{f, r}$.
\ENDFOR
\end{algorithmic}
\end{algorithm}
\subsection{Overview} In the standard FL scheme \cite{mc+17}, the server orchestrates the whole learning process across participating clients with iterative communication of model parameters. Since our method is essentially a model mixture-based PFL method, we need two models same in structure per client (one for a federated model, the other for a local model), of which initialization is different from each other. As a result, the client's local models are trained using different random initializations than its federated model counterpart. Note that different initialization is not common in FL \cite{mc+17}; however, this is intended behavior in our scheme for promoting construction of the connected low-loss subspace between a federated model and a local model.  
\subsection{Notations} We first define arguments required for the federated optimization as follows: a total number of communication rounds ($\mathrm{R}$), a personalization start round ($\mathrm{L}$) a local batch size ($\mathrm{B}$), a number of local epochs ($\mathrm{E}$), a total number of clients ($\mathrm{K}$), a fraction of clients selected at each round ($\mathrm{C}$), and a local learning rate ($\eta$). Following that, three hyperparameters are required for the optimization of our proposed method: a mixing constant $\lambda$ sampled from the uniform distribution $\text{Unif}(0,1)$, a constant $\mu$ for proximity regularization on the federated model $\mathbf{w}^f$ so that it is not distant from the global model $\mathbf{w}^g$; and a constant $\nu$ for inducing \textit{connectivity} along a subspace between local and federated models ($\mathbf{w}^l$ \& $\mathbf{w}^f$).

\subsection{Problem Statement} Consider that each client $c_i$ ($i\in[\mathrm{K}]$) has its own dataset $\mathcal{D}_i=\{(\mathbf{x}_i,\mathbf{y}_i)\}$ of size $n_i$, as well as a model $\mathcal{W}_i$. We assume that datasets are non-IID across clients in PFL, which means that each dataset $\mathcal{D}_i$ is sampled independently from a corresponding distribution $\mathcal{P}_i$ on both input and label spaces $\mathcal{Z}=\mathcal{X}\times\mathcal{Y}$. A hypothesis $h\in\mathcal{H}$ can be learned by the objective function $l:\mathcal{H}\times\mathcal{Z}\rightarrow\mathbb{R}^{+}$ in the form of $l(h(\mathbf{x};\mathcal{W}),\mathbf{y})$. We denote the expected loss as $\mathcal{L}_{\mathcal{P}_i}(h_i, \mathcal{W}_i)=\mathbb{E}_{(\mathbf{x}_i,\mathbf{y}_i)\sim\mathcal{P}_i}[l(h_i(\mathbf{x}_i;\mathcal{W}_i),\mathbf{y}_i)]$, and the empirical loss as $\hat{\mathcal{L}}_{\mathcal{D}_i}(h_i, \mathcal{W}_i)=\frac{1}{n_i}\sum_{j=1}^{n_i}l_i(h_i(\mathbf{x}_i^j;\mathcal{W}_i),\mathbf{y}_i^j)$; a regularization term $\Omega(\mathcal{W}_i)$ can also be incorporated here. Then, the global objective of PFL is to optimize (1).

\vspace{-2mm}
\begin{align*}\numberthis
\min_{h_i\in\mathcal{H}}\frac{1}{\mathrm{K}}\sum_{i=1}^{\mathrm{K}}\mathcal{L}_{\mathcal{P}_i}(h_i, \mathcal{W}_i)
\end{align*}
We can minimize this global objective through the empirical loss minimization (2).
\begin{align*}\numberthis
&\min_{\mathcal{W}_1,...,\mathcal{W}_\mathrm{K}}\frac{1}{\mathrm{K}}\sum_{i=1}^{\mathrm{K}}\hat{\mathcal{L}}_{\mathcal{D}_i}(h_i, \mathcal{W}_i) \\
&=\frac{1}{\mathrm{K}}\sum_{i=1}^{\mathrm{K}}\Big\{{\frac{1}{n_i}\sum_{j=1}^{n_i} {l}_i(h_i(\mathbf{x}_i^j;\mathcal{W}_i),\mathbf{y}_i^j)+\Omega(\mathcal{W}_i)}\Big\}
\end{align*} 
A model mixture-based PFL assumes each client $c_i$ has a set of paired parameters $\mathbf{w}_i^{f}, \mathbf{w}_i^{l}\subseteq\mathcal{W}_i$. Note that each of which is previously defined as a federated model and a local model. 
Both of them are grouped as a client model $\mathcal{W}_i=G(\mathbf{w}_i^{f}, \mathbf{w}_i^{l})$ with the grouping operator $G$. This operator can be a concatenation (e.g., stacking semantically separated layers like feature extractor or classifier~\cite{fedrep, FedPer, lgfedavg}): $G(\mathbf{w}_i^{f}, \mathbf{w}_i^{l})=\text{Concat}(\mathbf{w}_i^{f}, \mathbf{w}_i^{l})$, a simple enumeration of parameters having the same structure~\cite{ditto, pfedme}: $G(\mathbf{w}_i^{f}, \mathbf{w}_i^{l})=\{\mathbf{w}_i^{f}, \mathbf{w}_i^{l}\}$, or a convex combination~\cite{apfl,mansour+20,l2sgd} given a constant $\lambda\in[0,1]$: $G(\mathbf{w}_i^{f}, \mathbf{w}_i^{l};\lambda)=(1-\lambda)\mathbf{w}_i^{f}+\lambda\mathbf{w}_i^{l}$. 

\subsection{Local Update} In \texttt{SuPerFed}, we suppose both models $\mathbf{w}_i^{f}, \mathbf{w}_i^{l}$ have the same structure and consider $G$ to be a function for constructing a convex combination. We will write $\lambda$ as an explicit input to the client model with a slight abuse of notation: $\mathcal{W}_i(\lambda):=G(\mathbf{w}_i^{f}, \mathbf{w}_i^{l};\lambda)=(1-\lambda)\mathbf{w}_i^{f}+\lambda\mathbf{w}_i^{l}$.
The main goal of the local update in \texttt{SuPerFed} is to find a flat wide minima between the local model and the federated model. In the initial round, each client receives a global model transmitted from the server, sets it as a federated model, and copies the structure to set a local model. As different initialization is required, clients initialize their local model again with its fresh random seed.
During the local update, a client mixes two models using $\lambda\sim\text{Unif}(0,1)$ as $\mathcal{W}_i(\lambda)=(1-\lambda)\mathbf{w}_i^{f}+\lambda\mathbf{w}_i^{l}$, each time with a fresh mini-batch including $(\mathbf{x}, \mathbf{y})$ sampled from local dataset $\mathcal{D}$. The local model and the federated model are reduced to be interpolated diversely during local training. Finally, it is optimized with a loss function $l(\cdot)$ specific to the task, for example, cross-entropy, with additional regularization $\Omega(\cdot)$ that is critical to induce the \textit{connectivity}.

\medskip
\begin{Remark}~Connectivity~\cite{frankle20, nnsubspaces}. There exists a low-loss subspace spanned by~$\mathcal{W}(\lambda)$ between two independently trained (i.e., trained from different random initializations) deep networks $\mathbf{w}_1 \text{ and } \mathbf{w}_2$, in the form of a linear combination formed by $\lambda$: $\mathcal{W}(\lambda) = (1 - \lambda)\mathbf{w}_1 + \lambda\mathbf{w}_2, \lambda\in[0, 1]$.
\end{Remark}

\medskip
\noindent
The local objective is
\begin{align*}\numberthis
\min_{\mathcal{W}_i}\hat{\mathcal{L}}_{\mathcal{D}}(h, \mathcal{W}_i(\lambda))
\end{align*}
, where $\hat{\mathcal{L}}_{\mathcal{D}}(h, \mathcal{W}_i(\lambda))$ is denoted as:
\begin{align*}\numberthis
\hat{\mathcal{L}}_{\mathcal{D}}(h, &\mathcal{W}_i(\lambda)) \\
&=\frac{1}{\vert\mathcal{D}\vert}\sum_{(\mathbf{x}, \mathbf{y})\in\mathcal{D}} l(h(\mathbf{x};\mathcal{W}_i(\lambda)),\mathbf{y})+\Omega(\mathcal{W}_i(\lambda)) \\
&=\frac{1}{\vert\mathcal{D}\vert}\sum_{(\mathbf{x}, \mathbf{y})\in\mathcal{D}}l(h(\mathbf{x};\mathcal{W}_i(\lambda)),\mathbf{y}) +\mu\left\Vert\mathbf{w}^{f}_{i}-\mathbf{w}^g\right\Vert_2^2 \\
&+\nu\cos ^{2}\left(\mathbf{w}^{f}_{i}, \mathbf{w}^{l}_{i}\right)
\end{align*}
Note that the global model $\mathbf{w}^g$ is fixed here for the proximity regularization of $\mathbf{w}^{f}_{i}$. 
Denote the local objective (3) simply as $\hat{\mathcal{L}}$. Then, the update of each endpoint $\mathbf{w}^{f}_{i}$ and $\mathbf{w}^{l}_{i}$ is done using the following estimates. Note that two endpoints (i.e., $\mathcal{W}_i(0)=\mathbf{w}^{f}_{i}$ and $\mathcal{W}_i(1)=\mathbf{w}^{l}_{i}$) can be jointly updated at the same time in a single training run. 
\begin{align*}\numberthis
\frac{\partial \hat{\mathcal{L}}}{\partial \mathbf{w}^{f}_{i}}=(1-\lambda)\frac{\partial l}{\partial \mathcal{W}_i(\lambda)}+\mu\frac{\partial \left\Vert\mathbf{w}^{f}_{i}-\mathbf{w}^g\right\Vert_2^2}{\partial \mathbf{w}^{f}_{i}}+\nu \frac{\partial \cos ^{2}\left(\mathbf{w}^{f}_{i}, \mathbf{w}^{l}_{i}\right)}{\partial \mathbf{w}^{f}_{i}}
\end{align*}

\begin{align*}\numberthis
\frac{\partial \hat{\mathcal{L}}}{\partial \mathbf{w}^{l}_{i}}=\lambda\frac{\partial l}{\partial \mathcal{W}_i(\lambda)}+\nu \frac{\partial \cos ^{2}\left(\mathbf{w}^{f}_{i}, \mathbf{w}^{l}_{i}\right)}{\partial \mathbf{w}^{l}_{i}}
\end{align*}

\subsection{Regularization}
The regularization term $\cos(\cdot,\cdot)$ stands for cosine similarity defined as:  $\cos(\mathbf{w}^{f}_{i}, \mathbf{w}^{l}_{i}) = {\langle\mathbf{w}^{f}_{i}, \mathbf{w}^{l}_{i}\rangle} / ({\left\Vert\mathbf{w}^{f}_{i}\right\Vert\left\Vert\mathbf{w}^{l}_{i}\right\Vert})$. This aims to induce \textit{connectivity} between the local model and the federated model, thereby allowing the connected subspace between them to contain weights yielding high accuracy for the given task. Its magnitude is controlled by the constant $\nu$. Since merely applying the mixing strategy between the local and the federated models has little benefit \cite{nnsubspaces}, it is required to set $\nu>0$. In \cite{fort2019deep, nnsubspaces}, it is observed that weights of independently trained deep networks show dissimilarity in terms of cosine similarity, which has functional diversity as a result. Moreover, in \cite{ortho2,ortho3}, inducing orthogonality (i.e., forcing cosine similarity to zero) between weights prevents deep networks from learning redundant features given their learning capacity. In this context, we expect the local model of each client $\mathbf{w}^{l}_{i}$ to learn client-specific knowledge while the federated model $\mathbf{w}^{f}_{i}$ learns client-agnostic knowledge in a complementary manner and is harmoniously combined with the local model. 
The other regularization term adjusted by $\mu$ is L2-norm which controls the proximity of the update of the federated model $\mathbf{w}^{f}_{i}$ from the global model $\mathbf{w}^{g}$~\cite{fedprox}. By doing so, we expect each local update of the federated model not to be derailed from the global model, which prevents divergence of the aggregated global model during the process of FL. Note that an inordinate degree of proximity regularization can hinder the local model's adaptation and thereby cause the divergence of a global model~\cite{fedprox, pfedme}. Thus, a moderate degree of $\mu$ is required. 
Note that our method is designed to be reduced to the existing FL methods by adjusting hyperparameters, and thus it can be guaranteed to have at least their performance. In detail, when fixing $\lambda=0, \nu=0, \mu=0$, the objective of \texttt{SuPerFed} is reduced to \texttt{FedAvg}~\cite{mc+17}, when $\lambda=0, \nu=0, \mu>0$, it is equivalent to \texttt{FedProx}~\cite{fedprox}.

\subsection{Communication with the Server} At each round, the server samples a fraction of clients at every round (suppose $N$ clients are sampled), and requests a local update to each client. Only the federated model $\textbf{w}_i^{f}$ and the amounts of consumed training data per client $n_i$ are transferred to the server after the local update is completed in parallel across selected clients. The server then aggregates them to create an updated global model $\mathbf{w}^g$ as $\mathbf{w}^{g}\leftarrow\frac{1}{\sum_{i=1}^{N}{n_i}}\sum_{i=1}^{N}{n_i}\mathbf{w}_i^{f}$.
In the next round, only the updated global model $\mathbf{w}^g$ is transmitted to another group of selected clients, and it is set to be a new federated model at each client: $\mathbf{w}_i^f\leftarrow\mathbf{w}^g$. It is worth noting that the communication cost remains constant as with the single model-based FL methods~\cite{mc+17, fedprox}. See Algorithm~\ref{alg:SuPerFed} for understanding the whole process of \texttt{SuPerFed}.

\subsection{Variation in Mixing Method} Until now, we only considered the setting of applying $\lambda$ identically to the whole layers of $\mathbf{w}^f$ and $\mathbf{w}^l$. We name this setting \textit{Model Mixing}, in short, \texttt{MM}. On the one hand, it is also possible to sample different $\lambda$ for weights of different layers, i.e., mixing two models in a layer-wise manner. We also adopt this setting and name it \textit{Layer Mixing}, in short, \texttt{LM}.

\begin{table}[!htb]
\centering
\caption{Experimental results on the \textit{pathological non-IID setting} (MNIST and CIFAR10 datasets) compared with other FL and PFL methods. The top-1 accuracy is reported with a standard deviation.}
\label{table1}
\resizebox{\columnwidth}{!}{%
\begin{tabular}{l|lll|lll}
\hline
\multicolumn{1}{c|}{Dataset} & \multicolumn{3}{c|}{MNIST} & \multicolumn{3}{c}{CIFAR10} \\ \hline
\multicolumn{1}{c|}{\# clients} & \multicolumn{1}{c}{50} & \multicolumn{1}{c}{100} & \multicolumn{1}{c|}{500} & \multicolumn{1}{c}{50} & \multicolumn{1}{c}{100} & \multicolumn{1}{c}{500} \\
\multicolumn{1}{c|}{\# samples} & \multicolumn{1}{c}{960} & \multicolumn{1}{c}{480} & \multicolumn{1}{c|}{96} & \multicolumn{1}{c}{800} & \multicolumn{1}{c}{400} & \multicolumn{1}{c}{80} \\ \hline
FedAvg & $95.69\pm2.39$ & $89.78\pm11.30$ & $96.04\pm4.74$ & $43.09\pm24.56$ & $36.19\pm29.54$ & $47.90\pm25.05$ \\
FedProx & $95.13\pm2.67$ & $93.25\pm6.12$ & $96.50\pm4.52$ & $49.01\pm19.87$ & $38.56\pm28.11$ & $48.60\pm25.71$ \\
SCAFFOLD & $95.50\pm2.71$ & $90.58\pm10.13$ & $96.60\pm4.26$ & $43.81\pm24.30$ & $36.31\pm29.42$ & $40.27\pm26.90$ \\ \hline
LG-FedAvg & $98.21\pm1.28$ & $97.52\pm2.11$ & $96.05\pm5.02$ & $89.03\pm4.53$ & $70.25\pm35.66$ & $78.52\pm11.22$ \\
FedPer & $99.23\pm0.66$ & $99.14\pm0.93$ & $98.67\pm2.61$ & $89.10\pm5.41$ & $87.99\pm5.70$ & $82.35\pm9.85$ \\
APFL & $99.40\pm0.58$ & $99.19\pm0.92$ & $98.98\pm2.22$ & $92.83\pm3.47$ & $91.73\pm4.61$ & $87.38\pm9.39$ \\
pFedMe & $81.10\pm8.52$ & $82.48\pm7.62$ & $81.96\pm12.28$ & $92.97\pm3.07$ & $92.07\pm5.05$ & $88.30\pm8.53$ \\
Ditto & $97.07\pm1.38$ & $97.13\pm2.06$ & $97.20\pm3.72$ & $85.53\pm6.22$ & $83.01\pm5.62$ & $84.45\pm10.67$ \\
FedRep & $99.11\pm0.63$ & $99.04\pm1.02$ & $97.94\pm3.37$ & $82.00\pm5.41$ & $81.27\pm7.90$ & $80.66\pm11.00$ \\ \hline
SuPerFed-MM & $99.45\pm0.46$ & $\mathbf{99.38\pm0.93}$ & $\mathbf{99.24\pm2.12}$ & $\mathbf{94.05\pm3.18}$ & $\mathbf{93.25\pm3.80}$ & $\mathbf{90.81\pm9.35}$ \\
SuPerFed-LM & $\mathbf{99.48\pm0.54}$ & $99.31\pm1.09$ & $98.83\pm3.02$ & $93.88\pm3.55$ & $93.20\pm4.19$ & $89.63\pm11.11$ \\ \hline
\end{tabular}%
}
\end{table}

\begin{table}[!htb]
\centering
\caption{Experimental results on the \textit{Dirichlet distribution-based non-IID setting} (CIFAR100 and TinyImageNet) compared with other FL and PFL methods. The top-5 accuracy is reported with a standard deviation.}
\label{table2}
\resizebox{\columnwidth}{!}{%
\begin{tabular}{l|lll|lll}
\hline
\multicolumn{1}{c|}{Dataset} & \multicolumn{3}{c|}{CIFAR100} & \multicolumn{3}{c}{TinyImageNet} \\ \hline
\multicolumn{1}{c|}{\# clients} & \multicolumn{3}{c|}{100} & \multicolumn{3}{c}{200} \\
\multicolumn{1}{c|}{concentration ($\alpha$)} & \multicolumn{1}{c}{1} & \multicolumn{1}{c}{10} & \multicolumn{1}{c|}{100} & \multicolumn{1}{c}{1} & \multicolumn{1}{c}{10} & \multicolumn{1}{c}{100} \\ \hline
FedAvg & $58.12\pm7.06$ & $59.04\pm7.19$ & $58.49\pm5.27$ & $46.61\pm5.64$ & $48.90\pm5.50$ & $48.90\pm5.40$ \\
FedProx & $57.71\pm6.79$ & $58.24\pm5.94$ & $58.75\pm5.56$ & $47.37\pm5.94$ & $47.73\pm5.94$ & $48.97\pm5.02$ \\
SCAFFOLD & $51.16\pm6.79$ & $51.40\pm5.22$ & $52.90\pm4.89$ & $46.54\pm5.49$ & $48.77\pm5.49$ & $48.27\pm5.32$ \\ \hline
LG-FedAvg & $28.88\pm5.64$ & $21.25\pm4.64$ & $20.05\pm4.61$ & $14.70\pm3.84$ & $9.86\pm3.13$ & $9.25\pm2.89$ \\
FedPer & $46.78\pm7.63$ & $35.73\pm6.80$ & $35.52\pm6.58$ & $21.90\pm4.71$ & $11.10\pm3.19$ & $9.63\pm3.12$ \\
APFL & $61.13\pm6.86$ & $56.90\pm7.05$ & $55.43\pm5.45$ & $41.98\pm5.94$ & $34.74\pm5.14$ & $34.23\pm5.07$ \\
pFedMe & $19.00\pm5.37$ & $17.94\pm4.72$ & $18.28\pm3.41$ & $6.05\pm2.84$ & $8.01\pm2.92$ & $7.69\pm2.41$ \\
Ditto & $60.04\pm6.82$ & $58.55\pm7.12$ & $58.73\pm5.39$ & $46.36\pm5.44$ & $43.84\pm5.44$ & $43.11\pm5.35$ \\
FedRep & $38.49\pm6.65$ & $26.61\pm5.20$ & $24.50\pm4.21$ & $18.67\pm4.66$ & $9.23\pm2.84$ & $8.09\pm2.83$ \\ \hline
SuPerFed-MM & $60.14\pm6.24$ & $58.32\pm6.25$ & $\mathbf{59.08\pm5.12}$ & $\mathbf{50.07\pm5.73}$ & $\mathbf{49.86\pm5.03}$ & $\mathbf{49.73\pm4.84}$ \\
SuPerFed-LM & $\mathbf{62.50\pm6.34}$ & $\mathbf{61.64\pm6.23}$ & $59.05\pm5.59$ & $47.28\pm5.19$ & $48.98\pm4.79$ & $49.29\pm4.82$ \\ \hline
\end{tabular}%
}
\end{table}

\begin{table}[!htb]
\centering
\caption{Experimental results on the \textit{realistic non-IID setting} (FEMNIST and Shakespeare) compared with other FL and PFL methods. The top-1 accuracy is reported with a standard deviation.}
\label{table3}
\resizebox{0.8\columnwidth}{!}{%
\begin{tabular}{l|ll|ll}
\hline
\multicolumn{1}{c|}{Dataset} & \multicolumn{2}{c|}{FEMNIST} & \multicolumn{2}{c}{Shakespeare} \\ \hline
\multicolumn{1}{c|}{\# clients} & \multicolumn{2}{c|}{730} & \multicolumn{2}{c}{660} \\
\multicolumn{1}{c|}{Accuracy} & \multicolumn{1}{c}{\begin{tabular}[c]{@{}c@{}}Top-1\end{tabular}} & \multicolumn{1}{c|}{\begin{tabular}[c]{@{}c@{}}Top-5\end{tabular}} & \multicolumn{1}{c}{\begin{tabular}[c]{@{}c@{}}Top-1\end{tabular}} & \multicolumn{1}{c}{\begin{tabular}[c]{@{}c@{}}Top-5\end{tabular}} \\ \hline
FedAvg & $80.12\pm12.01$ & $98.74\pm2.97$ & $50.90\pm7.85$ & $80.15\pm7.87$ \\
FedProx & $80.23\pm11.88$ & $98.73\pm2.94$ & $51.33\pm7.54$ & $80.31\pm6.95$ \\
SCAFFOLD & $80.03\pm11.78$ & $98.85\pm2.77$ & $50.76\pm8.01$ & $80.43\pm7.09$ \\ \hline
LG-FedAvg & $50.84\pm20.97$ & $75.11\pm21.49$ & $33.88\pm10.28$ & $62.84\pm13.16$ \\
FedPer & $73.79\pm14.10$ & $86.39\pm14.70$ & $45.82\pm8.10$ & $75.68\pm9.25$ \\
APFL & $84.85\pm8.83$ & $98.83\pm2.73$ & $54.08\pm8.31$ & $83.32\pm6.22$ \\
pFedMe & $5.98\pm4.55$ & $24.64\pm9.43$ & $32.29\pm6.64$ & $63.12\pm8.00$ \\
Ditto & $64.61\pm31.49$ & $81.14\pm28.56$ & $49.04\pm10.22$ & $78.14\pm12.61$ \\
FedRep & $59.27\pm15.72$ & $70.42\pm15.82$ & $38.15\pm9.54$ & $68.65\pm12.50$ \\ \hline
SuPerFed-MM & $\mathbf{85.20\pm8.40}$ & $\mathbf{99.16\pm2.13}$ & $54.52\pm7.54$ & $\mathbf{84.27\pm6.00}$ \\
SuPerFed-LM & $83.36\pm9.61$ & $98.81\pm2.58$ & $\mathbf{54.52\pm7.54}$ & $83.97\pm5.72$ \\ \hline
\end{tabular}%
}
\end{table}

\begin{table*}[]
\centering
\caption{Experimental results on the label noise simulation with \textit{Dirichlet distirubiotn-based non-IID setting} ($\alpha=100$) on (MNIST and CIFAR10) compared with other FL and PFL methods. Per each cell, expected calibration error (ECE), maximum calibration error (MCE), and the best averaged top-1 accuracy (in parentheses) are enumerated from top to bottom. Note that the lower ECE and MCE, the better the model calibration is.}
\label{table4}
\resizebox{0.9\textwidth}{!}{%
\begin{tabular}{l|cccc|cccc}
\hline
\multicolumn{1}{c|}{Dataset} & \multicolumn{4}{c|}{MNIST} & \multicolumn{4}{c}{CIFAR10} \\ \hline
\multicolumn{1}{c|}{Noise type} & \multicolumn{2}{c|}{pair} & \multicolumn{2}{c|}{symmetric} & \multicolumn{2}{c|}{pair} & \multicolumn{2}{c}{symmetric} \\ \hline
\multicolumn{1}{c|}{Noise ratio} & 0.1 & \multicolumn{1}{c|}{0.4} & 0.2 & 0.6 & 0.1 & \multicolumn{1}{c|}{0.4} & 0.2 & 0.6 \\ \hline
FedAvg & \begin{tabular}[c]{@{}c@{}}$0.17\pm0.03$\\ $0.58\pm0.08$\\ $(82.40\pm3.31)$\end{tabular} & \multicolumn{1}{c|}{\begin{tabular}[c]{@{}c@{}}$0.38\pm0.03$\\ $\mathbf{0.67\pm0.04}$\\ $(41.01\pm4.64)$\end{tabular}} & \begin{tabular}[c]{@{}c@{}}$0.29\pm0.04$\\ $0.66\pm0.07$\\ $(66.94\pm4.27)$\end{tabular} & \begin{tabular}[c]{@{}c@{}}$0.42\pm0.04$\\ $0.75\pm0.05$\\ $(49.52\pm5.42)$\end{tabular} & \begin{tabular}[c]{@{}c@{}}$0.46\pm0.04$\\ $0.80\pm0.06$\\ $(45.08\pm5.61)$\end{tabular} & \multicolumn{1}{c|}{\begin{tabular}[c]{@{}c@{}}$0.57\pm0.04$\\ $0.87\pm0.04$\\ $(20.90\pm4.66)$\end{tabular}} & \begin{tabular}[c]{@{}c@{}}$0.52\pm0.05$\\ $0.81\pm0.04$\\ $(38.62\pm5.28)$\end{tabular} & \begin{tabular}[c]{@{}c@{}}$0.59\pm0.05$\\ $0.84\pm0.04$\\ $(30.18\pm5.53)$\end{tabular} \\ \hline
FedProx & \begin{tabular}[c]{@{}c@{}}$0.17\pm0.03$\\ $0.58\pm0.07$\\ $(82.05\pm3.98)$\end{tabular} & \multicolumn{1}{c|}{\begin{tabular}[c]{@{}c@{}}$0.38\pm0.03$\\ $0.78\pm0.05$\\ $(41.39\pm4.61)$\end{tabular}} & \begin{tabular}[c]{@{}c@{}}$0.29\pm0.03$\\ $0.66\pm0.07$\\ $(67.15\pm4.60)$\end{tabular} & \begin{tabular}[c]{@{}c@{}}$0.42\pm0.05$\\ $0.74\pm0.05$\\ $(49.98\pm5.57)$\end{tabular} & \begin{tabular}[c]{@{}c@{}}$0.47\pm0.05$\\ $0.80\pm0.05$\\ $(44.31\pm6.20)$\end{tabular} & \multicolumn{1}{c|}{\begin{tabular}[c]{@{}c@{}}$0.70\pm0.05$\\ $0.87\pm0.04$\\ $(21.58\pm4.62)$\end{tabular}} & \begin{tabular}[c]{@{}c@{}}$0.53\pm0.05$\\ $0.81\pm0.05$\\ $(36.91\pm5.68)$\end{tabular} & \begin{tabular}[c]{@{}c@{}}$0.59\pm0.05$\\ $0.84\pm0.05$\\ $(29.50\pm6.11)$\end{tabular} \\ \hline
SCAFFOLD & \begin{tabular}[c]{@{}c@{}}$0.16\pm0.03$\\ $0.58\pm0.07$\\ $(60.86\pm4.09)$\end{tabular} & \multicolumn{1}{c|}{\begin{tabular}[c]{@{}c@{}}$0.45\pm0.04$\\ $0.77\pm0.04$\\ $(44.92\pm5.07)$\end{tabular}} & \begin{tabular}[c]{@{}c@{}}$0.29\pm0.04$\\ $\mathbf{0.59\pm0.08}$\\ $(70.51\pm4.25)$\end{tabular} & \begin{tabular}[c]{@{}c@{}}$0.44\pm0.04$\\ $0.73\pm0.05$\\ $(51.46\pm5.12)$\end{tabular} & \begin{tabular}[c]{@{}c@{}}$0.46\pm0.05$\\ $0.76\pm0.05$\\ $(47.54\pm5.64)$\end{tabular} & \multicolumn{1}{c|}{\begin{tabular}[c]{@{}c@{}}$0.65\pm0.04$\\ $0.86\pm0.03$\\ $(22.72\pm4.01)$\end{tabular}} & \begin{tabular}[c]{@{}c@{}}$0.53\pm0.04$\\ $0.79\pm0.04$\\ $(38.85\pm5.45)$\end{tabular} & \begin{tabular}[c]{@{}c@{}}$0.61\pm0.04$\\ $0.83\pm0.04$\\ $(30.18\pm4.63)$\end{tabular} \\ \hline
LG-FedAvg & \begin{tabular}[c]{@{}c@{}}$0.23\pm0.04$\\ $0.66\pm0.08$\\ $(73.65\pm5.32)$\end{tabular} & \multicolumn{1}{c|}{\begin{tabular}[c]{@{}c@{}}$0.50\pm0.05$\\ $0.81\pm0.04$\\ $(37.69\pm5.41)$\end{tabular}} & \begin{tabular}[c]{@{}c@{}}$0.34\pm0.04$\\ $0.71\pm0.07$\\ $(61.54\pm4.96)$\end{tabular} & \begin{tabular}[c]{@{}c@{}}$0.45\pm0.05$\\ $0.75\pm0.06$\\ $(47.79\pm5.02)$\end{tabular} & \begin{tabular}[c]{@{}c@{}}$0.59\pm0.06$\\ $0.83\pm0.05$\\ $(30.22\pm6.49)$\end{tabular} & \multicolumn{1}{c|}{\begin{tabular}[c]{@{}c@{}}$0.69\pm0.05$\\ $0.89\pm0.03$\\ $(17.44\pm4.66)$\end{tabular}} & \begin{tabular}[c]{@{}c@{}}$0.63\pm0.05$\\ $0.85\pm0.04$\\ $(25.68\pm5.18)$\end{tabular} & \begin{tabular}[c]{@{}c@{}}$0.66\pm0.05$\\ $0.87\pm0.03$\\ $(22.04\pm5.56)$\end{tabular} \\ \hline
FedPer & \begin{tabular}[c]{@{}c@{}}$0.17\pm0.03$\\ $0.57\pm0.08$\\ $(82.43\pm4.18)$\end{tabular} & \multicolumn{1}{c|}{\begin{tabular}[c]{@{}c@{}}$0.40\pm0.04$\\ $0.78\pm0.05$\\ $(40.75\pm5.49)$\end{tabular}} & \begin{tabular}[c]{@{}c@{}}$0.28\pm0.04$\\ $0.66\pm0.08$\\ $(68.44\pm5.63)$\end{tabular} & \begin{tabular}[c]{@{}c@{}}$0.40\pm0.04$\\ $0.73\pm0.07$\\ $(52.41\pm5.56)$\end{tabular} & \begin{tabular}[c]{@{}c@{}}$0.54\pm0.05$\\ $0.81\pm0.06$\\ $(39.81\pm6.02)$\end{tabular} & \multicolumn{1}{c|}{\begin{tabular}[c]{@{}c@{}}$0.70\pm0.05$\\ $0.87\pm0.04$\\ $(20.37\pm5.43)$\end{tabular}} & \begin{tabular}[c]{@{}c@{}}$0.60\pm0.06$\\ $0.82\pm0.05$\\ $(32.82\pm5.90)$\end{tabular} & \begin{tabular}[c]{@{}c@{}}$0.65\pm0.05$\\ $0.85\pm0.04$\\ $(26.82\pm5.38)$\end{tabular} \\ \hline
APFL & \begin{tabular}[c]{@{}c@{}}$0.18\pm0.03$\\ $0.60\pm0.07$\\ $(80.18\pm4.57)$\end{tabular} & \multicolumn{1}{c|}{\begin{tabular}[c]{@{}c@{}}$0.42\pm.0.05$\\ $0.78\pm0.06$\\ $(40.43\pm6.06)$\end{tabular}} & \begin{tabular}[c]{@{}c@{}}$0.28\pm0.05$\\ $0.67\pm0.06$\\ $(67.83\pm5.53)$\end{tabular} & \begin{tabular}[c]{@{}c@{}}$0.40\pm0.05$\\ $0.79\pm0.07$\\ $(52.15\pm5.63)$\end{tabular} & \begin{tabular}[c]{@{}c@{}}$0.45\pm0.06$\\ $0.78\pm0.06$\\ $(45.27\pm6.21)$\end{tabular} & \multicolumn{1}{c|}{\begin{tabular}[c]{@{}c@{}}$0.60\pm0.06$\\ $0.86\pm0.04$\\ $(23.42\pm5.49)$\end{tabular}} & \begin{tabular}[c]{@{}c@{}}$0.51\pm0.06$\\ $0.81\pm0.05$\\ $(37.85\pm5.84)$\end{tabular} & \begin{tabular}[c]{@{}c@{}}$0.56\pm0.06$\\ $0.83\pm0.05$\\ $(31.46\pm5.73)$\end{tabular} \\ \hline
pFedMe & \begin{tabular}[c]{@{}c@{}}$0.23\pm0.04$\\ $0.66\pm0.07$\\ $(72.05\pm0.05)$\end{tabular} & \multicolumn{1}{c|}{\begin{tabular}[c]{@{}c@{}}$0.46\pm0.04$\\ $0.80\pm.0.05$\\ $(37.89\pm5.77)$\end{tabular}} & \begin{tabular}[c]{@{}c@{}}$0.33\pm0.05$\\ $0.72\pm0.06$\\ $(59.80\pm5.94)$\end{tabular} & \begin{tabular}[c]{@{}c@{}}$0.44\pm0.04$\\ $0.76\pm0.05$\\ $(45.79\pm5.26)$\end{tabular} & \begin{tabular}[c]{@{}c@{}}$0.58\pm0.06$\\ $0.83\pm0.52$\\ $(29.62\pm6.37)$\end{tabular} & \multicolumn{1}{c|}{\begin{tabular}[c]{@{}c@{}}$0.66\pm0.04$\\ $0.88\pm0.03$\\ $(17.94\pm3.97)$\end{tabular}} & \begin{tabular}[c]{@{}c@{}}$0.60\pm0.05$\\ $0.85\pm0.04$\\ $(26.01\pm0.05)$\end{tabular} & \begin{tabular}[c]{@{}c@{}}$0.64\pm0.05$\\ $0.87\pm0.03$\\ $(21.36\pm4.42)$\end{tabular} \\ \hline
Ditto & \begin{tabular}[c]{@{}c@{}}$0.22\pm0.04$\\ $0.66\pm0.07$\\ $(72.39\pm5.25)$\end{tabular} & \multicolumn{1}{c|}{\begin{tabular}[c]{@{}c@{}}$0.42\pm0.05$\\ $0.79\pm0.05$\\ $(38.20\pm6.13)$\end{tabular}} & \begin{tabular}[c]{@{}c@{}}$0.31\pm0.04$\\ $0.69\pm0.07$\\ $(60.62\pm5.67)$\end{tabular} & \begin{tabular}[c]{@{}c@{}}$0.41\pm0.04$\\ $0.77\pm0.05$\\ $(45.11\pm4.95)$\end{tabular} & \begin{tabular}[c]{@{}c@{}}$0.54\pm0.05$\\ $0.84\pm0.05$\\ $(29.41\pm5.15)$\end{tabular} & \multicolumn{1}{c|}{\begin{tabular}[c]{@{}c@{}}$0.60\pm0.06$\\ $0.88\pm0.04$\\ $(18.17\pm4.43)$\end{tabular}} & \begin{tabular}[c]{@{}c@{}}$0.56\pm0.07$\\ $0.85\pm0.04$\\ $(26.39\pm5.83)$\end{tabular} & \begin{tabular}[c]{@{}c@{}}$0.58\pm0.06$\\ $0.87\pm0.04$\\ $(21.72\pm4.90)$\end{tabular} \\ \hline
FedRep & \begin{tabular}[c]{@{}c@{}}$0.20\pm0.04$\\ $0.62\pm0.08$\\ $(77.95\pm4.65)$\end{tabular} & \multicolumn{1}{c|}{\begin{tabular}[c]{@{}c@{}}$0.53\pm0.05$\\ $0.81\pm0.05$\\ $(35.88\pm5.15)$\end{tabular}} & \begin{tabular}[c]{@{}c@{}}$0.30\pm0.05$\\ $0.68\pm0.08$\\ $(66.58\pm5.60)$\end{tabular} & \begin{tabular}[c]{@{}c@{}}$0.43\pm0.05$\\ $0.74\pm0.06$\\ $(51.30\pm5.27)$\end{tabular} & \begin{tabular}[c]{@{}c@{}}$0.62\pm0.05$\\ $0.82\pm0.05$\\ $(33.10\pm5.23)$\end{tabular} & \multicolumn{1}{c|}{\begin{tabular}[c]{@{}c@{}}$0.76\pm0.04$\\ $0.87\pm0.03$\\ $(17.80\pm4.43)$\end{tabular}} & \begin{tabular}[c]{@{}c@{}}$0.66\pm0.05$\\ $0.83\pm0.05$\\ $(29.22\pm5.67)$\end{tabular} & \begin{tabular}[c]{@{}c@{}}$0.71\pm0.05$\\ $0.85\pm0.04$\\ $(22.94\pm5.20)$\end{tabular} \\ \hline
SuPerFed-MM & \begin{tabular}[c]{@{}c@{}}$0.16\pm0.03$\\ $0.53\pm0.07$\\ $(83.67\pm3.51)$\end{tabular} & \multicolumn{1}{c|}{\begin{tabular}[c]{@{}c@{}}$\mathbf{0.28\pm0.04}$\\$0.69\pm0.06$\\ $(46.41\pm5.14)$\end{tabular}} & \begin{tabular}[c]{@{}c@{}}$\mathbf{0.27\pm0.03}$\\ $0.64\pm0.07$\\ $(\mathbf{71.99\pm5.01})$\end{tabular} & \begin{tabular}[c]{@{}c@{}}$\mathbf{0.30\pm0.04}$\\ $\mathbf{0.69\pm0.08}$\\ $(\mathbf{56.07\pm4.32})$\end{tabular} & \begin{tabular}[c]{@{}c@{}}$\mathbf{0.28\pm0.06}$\\ $0.69\pm0.07$\\ $(48.79\pm5.73)$\end{tabular} & \multicolumn{1}{c|}{\begin{tabular}[c]{@{}c@{}}$\mathbf{0.27\pm0.05}$\\ $\mathbf{0.63\pm0.09}$\\ $(26.64\pm4.34)$\end{tabular}} & \begin{tabular}[c]{@{}c@{}}$\mathbf{0.31\pm0.06}$\\ $0.71\pm0.10$\\ $42.66\pm5.61$\end{tabular} & \begin{tabular}[c]{@{}c@{}}$\mathbf{0.28\pm0.06}$\\ $\mathbf{0.68\pm0.10}$\\ $(\mathbf{35.74\pm5.21})$\end{tabular} \\ \hline
SuPerFed-LM & \begin{tabular}[c]{@{}c@{}}$\mathbf{0.14\pm0.03}$\\ $\mathbf{0.49\pm0.08}$\\ $(\mathbf{84.78\pm3.63})$\end{tabular} & \multicolumn{1}{c|}{\begin{tabular}[c]{@{}c@{}}$0.35\pm0.04$\\ $0.77\pm0.04$\\ $(\mathbf{46.82\pm5.39})$\end{tabular}} & \begin{tabular}[c]{@{}c@{}}$0.27\pm0.04$\\ $0.66\pm0.07$\\ $(69.23\pm5.25)$\end{tabular} & \begin{tabular}[c]{@{}c@{}}$0.32\pm0.04$\\ $0.72\pm0.06$\\ $(54.69\pm4.20)$\end{tabular} & \begin{tabular}[c]{@{}c@{}}$0.29\pm0.04$\\ $\mathbf{0.68\pm0.06}$\\ $(\mathbf{51.44\pm5.67})$\end{tabular} & \multicolumn{1}{c|}{\begin{tabular}[c]{@{}c@{}}$0.40\pm0.04$\\ $0.80\pm0.04$\\ $(\mathbf{28.51\pm4.66})$\end{tabular}} & \begin{tabular}[c]{@{}c@{}}$0.36\pm0.05$\\ $\mathbf{0.70\pm0.07}$\\ $(\mathbf{42.81\pm5.50})$\end{tabular} & \begin{tabular}[c]{@{}c@{}}$0.37\pm0.06$\\ $0.76\pm0.05$\\ $(34.10\pm5.18)$\end{tabular} \\ \hline
\end{tabular}%
}
\end{table*}

\begin{table}[]
\centering
\caption{Experiment on selecting $\mathrm{L}$, the starting round of personalization. The best averaged top-1 accuracy (top) and loss (bottom) among models that can be realized by $\lambda\in[0, 1]$ is reported with a standard deviation.}
\label{table5}
\resizebox{\columnwidth}{!}{%
\begin{tabular}{c|ccccc}
\hline
$\mathrm{L}/\mathrm{R}$ & 0.0 & 0.2 & 0.4 & 0.6 & 0.8 \\ \hline
SuPerFed-MM & \begin{tabular}[c]{@{}c@{}}$89.87\pm4.73$\\ $1.36\pm0.58$\end{tabular} & \begin{tabular}[c]{@{}c@{}}$92.67\pm4.04$\\ $\mathbf{0.72\pm0.40}$\end{tabular} & \begin{tabular}[c]{@{}c@{}}$\mathbf{92.69\pm3.86}$\\ $0.78\pm0.31$\end{tabular} & \begin{tabular}[c]{@{}c@{}}$92.50\pm4.14$\\ $0.74\pm0.88$\end{tabular} & \begin{tabular}[c]{@{}c@{}}$91.29\pm6.28$\\ $2.53\pm0.72$\end{tabular} \\ \hline
SuPerFed-LM & \begin{tabular}[c]{@{}c@{}}$91.55\pm4.26$\\ $1.43\pm0.81$\end{tabular} & \begin{tabular}[c]{@{}c@{}}$92.41\pm3.93$\\ $1.18\pm0.60$\end{tabular} & \begin{tabular}[c]{@{}c@{}}$\mathbf{92.69\pm3.66}$\\ $0.94\pm0.45$\end{tabular} & \begin{tabular}[c]{@{}c@{}}$92.60\pm3.86$\\ $\mathbf{0.63\pm0.33}$\end{tabular} & \begin{tabular}[c]{@{}c@{}}$90.59\pm9.93$\\ $6.28\pm1.76$\end{tabular} \\ \hline
\end{tabular}%
}
\vspace{-2mm}
\end{table}

\section{Experiments}
\subsection{Setup} We focus on two points in our experiments: (i) personalization performance in various non-IID scenarios, (ii) robustness to label noise through extensive benchmark datasets. Details of each experiment are provided in Appendix A. Throughout all experiments, if not specified, we set 5 clients to be sampled at every round and used stochastic gradient descent (SGD) optimization with a momentum of 0.9 and a weight decay factor of 0.0001. 
As \texttt{SuPerFed} is a model mixture-based PFL method, we compare its performance to other model mixture-based PFL methods: \texttt{FedPer}~\cite{FedPer}, \texttt{LG-FedAvg}~\cite{lgfedavg}, \texttt{APFL}~\cite{apfl}, \texttt{pFedMe}~\cite{pfedme}, \texttt{Ditto}~\cite{ditto}, \texttt{FedRep}~\cite{fedrep}, along with basic single-model based FL methods, \texttt{FedAvg}~\cite{mc+17} \texttt{FedProx}~\cite{fedprox} and \texttt{SCAFFOLD}~\cite{ka+19}.
We let the client randomly split their data into a training set and a test set with a fraction of 0.2 to estimate the performance of each FL algorithm on each client's test set with its own personalized model (if possible) with various metrics: top-1 accuracy, top-5 accuracy, expected calibration error (ECE~\cite{ece}), and maximum calibration error (MCE~\cite{mce}). Note that we evaluated all participating clients after the whole round of FL is finished. For each variation of the mixing method (i.e., \texttt{LM} \& \texttt{MM}) stated in section 3.7, we subdivide our method into two, \texttt{SuPerFed-MM} for \textit{Model Mixing}, and \texttt{SuPerFed-LM} for \textit{Layer Mixing}.

\subsection{Personalization} For the estimation of the performance of \texttt{SuPerFed} as PFL methods, we simulate three different non-IID scenarios. (i) a \textit{pathological non-IID} setting proposed by~\cite{mc+17}, which assumes most clients have samples from two classes for a multi-class classification task. (2) \textit{Dirichlet distribution}-based non-IID setting proposed by \cite{diri}, in which the Dirichlet distribution with its concentration parameter $\alpha$ determines the label distribution of each client. All clients have samples from only one class when using $\alpha\rightarrow0$, whereas $\alpha\rightarrow\infty$ divides samples into an identical distribution. (3) \textit{Realistic} non-IID setting proposed in \cite{leaf}, which provides several benchmark datasets for PFL.

\medskip 
\noindent\textbf{Pathological non-IID setting.} In this setting, we used two multi-class classification datasets, MNIST~\cite{mnist} and CIFAR10~\cite{cifar} and both have 10 classes. We used the two-layered fully-connected network for the MNIST dataset and the two-layered convolutional neural network (CNN) for the CIFAR10 dataset as proposed in~\cite{mc+17}. For each dataset, we set the number of total clients ($\mathrm{K}$=50, 100, 500) to check the scalability of the PFL methods. 
The results of the \textit{pathological non-IID setting} are shown in Table~\ref{table1}. It is notable that our proposed method beats most of the existing model mixture-based PFL methods with a small standard deviation.

\smallskip 
\noindent\textbf{Dirichlet distribution-based non-IID setting.} In this setting, we used other multi-class classification datasets to simulate a more challenging setting than the \textit{pathological non-IID setting}. We use CIFAR100\cite{cifar} and TinyImageNet\cite{tinyimagenet} datasets, having 100 classes and 200 classes each. We also selected other deep architectures, ResNet\cite{resnet} for CIFAR100 and MobileNet\cite{mobilenet} for TinyImageNet. For each dataset, we adjust the concentration parameter $\alpha={1, 10, 100}$ to control the degree of statistical heterogeneity across clients. It is a more natural choice for simulating statistical heterogeneity than \textit{pathological non-IID setting}, making it more challenging. Note that the smaller the $\alpha$, the more heterogeneous each client's data distribution is. Table~\ref{table2} presents the results of the \textit{Dirichlet distribution-based non-IID setting}. Both of our methods are less affected by the degree of statistical heterogeneity (i.e., non-IIDness) determined by $\alpha$.

\medskip
\noindent\textbf{Realistic non-IID setting.} We used FEMNIST and Shakespeare datasets in the LEAF benchmark~\cite{leaf}. As the purpose of these datasets is to simulate a realistic FL scenario, each dataset is naturally split for a certain data mining task. The selected two datasets are for multi-class classification (62 classes) and next character prediction (80 characters) given a sentence, respectively. We used VGG~\cite{vgg} for the FEMNIST dataset and networks with two stacked LSTM layers were used in~\cite{mc+17} for the Shakespeare dataset. The results of the \textit{realistic non-IID setting} are shown in Table~\ref{table3}. In this massively distributed setting, our methods showed a consistent gain in terms of personalization performance.

\medskip 
\noindent\textbf{Robustness to label noise.} In realistic FL services, it is possible that each client suffers from noisy labels. We anticipate that \texttt{SuPerFed} will benefit from the strength of ensemble learning because our method can be viewed as implicit ensemble of two models, $\mathbf{w}^f$ and $\mathbf{w}^l$. One of the strengths is robustness to the label noise, as each of the model components of the ensemble method can view diverse data distributions~\cite{ensemblenoise,ensemblenoise2}. In \texttt{SuPerFed}, the federated model and the local model are induced to be independent of each other (i.e., having low cosine similarity in the weight space). We assume each of them should inevitably learn from a different view of data distribution. To prove this, we adopted two representative methods to simulate the noisy labels: pair flipping~\cite{pair} and symmetric flipping~\cite{symm}, and their detailed descriptions are in Appendix B. We select the noise ratio from {0.1, 0.4} for the pair flipping label noise, and {0.2, 0.6} for the symmetric flipping label noise. Using this noise scheme, we obfuscate each client's training set labels while labels of test set are kept intact. The results are shown in Table~\ref{table4}.
We used MNIST and CIFAR10 datasets with a Dirichlet distribution-based non-IID setting in which the concentration parameter $\alpha=10.0$. Since the robustness of the label noise can be quantified through evaluating ECE~\cite{ece} and MCE~\cite{mce}, we introduced these two metrics additionally. Both ECE and MCE measure the consistency between the prediction accuracy and the prediction confidence (i.e., calibration), thus lower values are preferred. From all the extensive experiments, our method shows stable performance in terms of calibration (i.e., low ECE and MCE), and therefore overall performance is not degraded much.
\vspace{-1.5mm}

\begin{figure*}[h]
\centering
\includegraphics[width=0.9\textwidth]{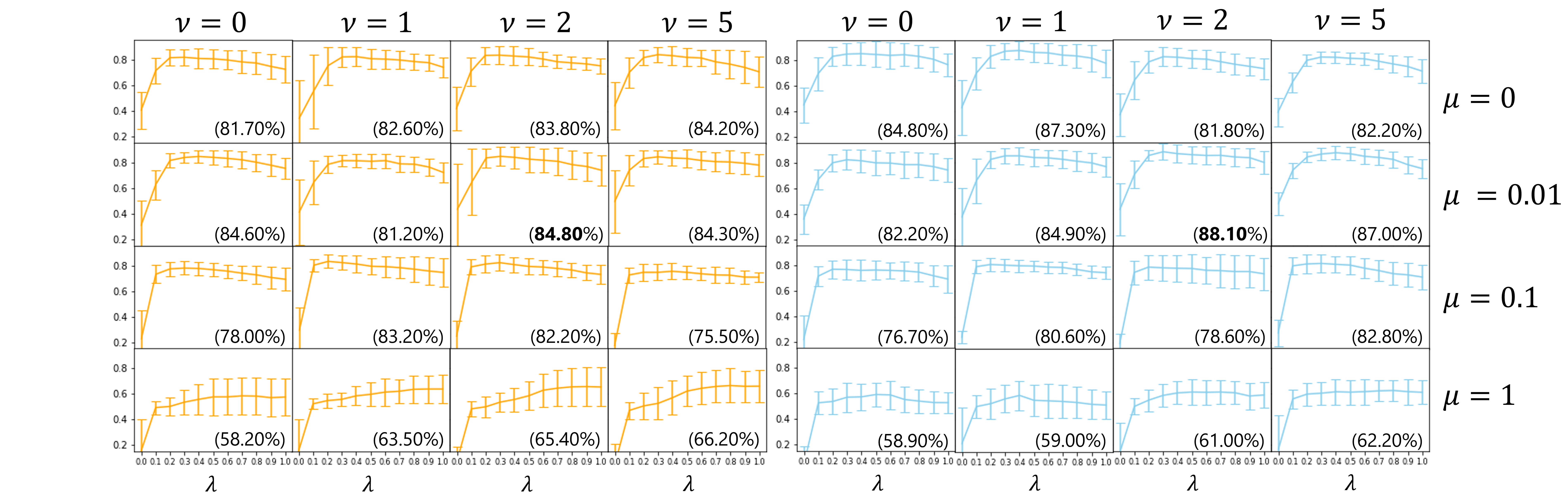}
\caption{The effects of hyperparameters $\nu$ and $\mu$. Left group of plots colored in orange is the performance of \texttt{SuPerFed-MM}, right group of plots colored in sky-blue is the performance of \texttt{SuPerFed-LM}. Each subplot's vertical axis represents accuracy, and the horizontal axis represents the range of possible $\lambda$ values from 0 to 1, with a 0.1 interval. Metrics on each subplot indicate the performance of the best averaged top-1 accuracy evaluated by \textit{local test set} of each client. The error bar indicates a standard deviation of each personalized model realized by different values of $\lambda\in[0,1]$. Note that each endpoint means a federated model ($\mathbf{w}^f$ when $\lambda=0$) or a local model ($\mathbf{w}^l$ when $\lambda=1$).}
\label{fig3}
\end{figure*}

\subsection{Ablation Studies} 
\noindent\textbf{Constants for regularization terms.} 
Our proposed method has two hyperparameters, $\nu$ and $\mu$. Each hyperparameter is supposed to do two things: (1) create a connected low-loss subspace between the federated model and the local model in clients by making the weights of the two models orthogonal; and (2) penalize local updates from going in the wrong direction, away from the direction of global updates.
We ran experiments with $\nu=\{0.0, 1.0, 2.0, 5.0\}$ and $\mu=\{0.0, 0.01, 0.1, 1.0\}$ (a total of 16 combinations) under the \textit{Model Mixing} and \textit{Layer Mixing} regimes, with $\mathrm{R}=100$ and $\mathrm{L}=\lfloor0.4\mathrm{R}\rfloor=40$, which yielded the best personalization performance on the CIFAR10 dataset with pathological non-IID Experimental results on these combinations are displayed in Figure~\ref{fig3}.

There are noticeable observations supporting the efficacy of our method. First, a learning strategy to find a connected low loss subspace between local and federated models yields high personalization performance. In most cases with non-zero $\nu$ and $\mu$, a combination of federated and local models (i.e., models realized by $\lambda\in(0,1)$) outperforms two single models from both endpoints (i.e., a federated model ($\lambda=0$) and a local model ($\lambda=1$). Second, a moderate degree of orthogonality regularization (adjusted by $\nu$) boosts the personalization performance of the local model.
Third, the proximity regularization restricts learning too much when $\mu$ is large, but often helps a federated model not to diverge when it is properly selected. Since this phenomenon has already been pointed out in ~\cite{fedprox, pfedme}, a careful selection of $\mu$ is required.

\smallskip
\noindent\textbf{Start round of personalization.} We conducted an experiment on selecting $\mathrm{L}$, which determines the initiation of mixing operation, in other words, sampling $\lambda\sim\text{Unif}(0, 1)$. As our proposed method can choose a mixing scheme from either \textit{Model Mixing} or \textit{Layer Mixing}, we designed separate experiments per scheme. We denote each scheme by \texttt{SuPerFed-MM} and \texttt{SuPerFed-LM}. By adjusting $\mathrm{L}$ by 0.1 increments from 0 to 0.9 of $\mathrm{R}$, we evaluated the top-1 accuracy of local models. This experiment was done only using the CIFAR10 dataset with $\mathrm{R}=200$, $\nu=2,\text{ and } \mu=0.01$. Experimental results are summarized in Table~\ref{table5}. 

We see that there are some differences according to the mixing scheme. In terms of the personalization performance, setting $\mathrm{L}/\mathrm{R}=0.4$ shows the best performance with a small standard deviation. This implies that allowing some rounds to a federated model to concentrate on learning global knowledge is a valid strategy. While $\mathrm{L}$ is a tunable hyperparameter, it can be adjusted for the purpose of FL. To derive a good global model, choosing a large $\mathrm{L}$ close to $\mathrm{R}$ is a valid choice, while in terms of personalization, it should be far from $\mathrm{R}$.

\section{Conclusion}
In this paper, we propose a simple and scalable personalized federated learning method \texttt{SuPerFed} under the scheme of the model mixture method. Its goal is to improve personalization performance while also providing robustness to label noise. In each client, the method aims to build a connected low-loss subspace between a local and a federated model, inspired by the special property of deep networks, \textit{connectivity}. Through this, two models are connected in a mutually cooperative manner so that each client has a personalized model with good performance and well-calibrated, which is robust to possible label noises in its dataset. With extensive experiments, we empirically demonstrated that \texttt{SuPerFed}, with its two variants in the mixing scheme, outperforms the existing model mixture-based PFL methods and three basic FL methods.

\begin{acks}
This work was supported by the National Research Foundation of Korea (NRF) Grant funded by the Korea Government (MSIT) under Grant No. 2020R1C1C1011063 and by Institute of Information and Communications Technology Planning and Evaluation (IITP) grant funded by the Korea Government (MSIT) (No. 2020-0-01336, Artificial Intelligence Graduate School support (UNIST)). This work is completed during internship at Kakao Enterprise.
\end{acks}

\bibliographystyle{ACM-Reference-Format}
\bibliography{bibfile}

\clearpage
\appendix
\setcounter{table}{0}
\renewcommand{\thetable}{A\arabic{table}}
\setcounter{figure}{0}
\renewcommand{\thefigure}{A\arabic{figure}}

\begin{table*}[!htbp]
\caption{Details of training configurations}
\label{tableA}
\centering 
\setcounter{table}{0}
\renewcommand{\thetable}{A\arabic{table}}
\begin{center}
\begin{tabular}{l|cc|cc|cc|cc}
\hline
\multicolumn{1}{c|}{Setting} & \multicolumn{2}{c|}{\begin{tabular}[c]{@{}c@{}}Pathological\\ non-IID\end{tabular}} & \multicolumn{2}{c|}{\begin{tabular}[c]{@{}c@{}}Dirichlet distribution-based\\ non-IID\end{tabular}} & \multicolumn{2}{c|}{\begin{tabular}[c]{@{}c@{}}Realistic\\ non-IID\end{tabular}} & \multicolumn{2}{c}{\begin{tabular}[c]{@{}c@{}}Label noise\\ (pair/symmetric)\end{tabular}} \\ \hline
\multicolumn{1}{c|}{Dataset} & MNIST & CIFAR10 & CIFAR100 & TinyImageNet & FEMNIST & Shakespeare & MNIST & CIFAR10 \\ \hline
$\mathrm{R}$ & 500 & 500 & 500 & 500 & 500 & 500 & 500 & 500 \\
$\mathrm{E}$ & 10 & 10 & 5 & 5 & 10 & 5 & 10 & 10 \\
$\mathrm{B}$ & 10 & 10 & 20 & 20 & 10 & 50 & 10 & 10 \\
$\mathrm{K}$ & 50,100,500 & 50,100,500 & 100 & 200 & 730 & 660 & 100 & 100 \\
$\mathrm{\eta}$ & 0.01 & 0.01 & 0.01 & 0.02 & 0.01 & 0.8 & 0.01 & 0.01 \\
Model & TwoNN & TwoCNN & ResNet & MobileNet & VGG & StackedLSTM & TwoNN & TwoCNN \\ \hline
\end{tabular}
\end{center}
\end{table*}

\begin{figure*}[!htbp]
\centering
\includegraphics[width=1.0\textwidth]{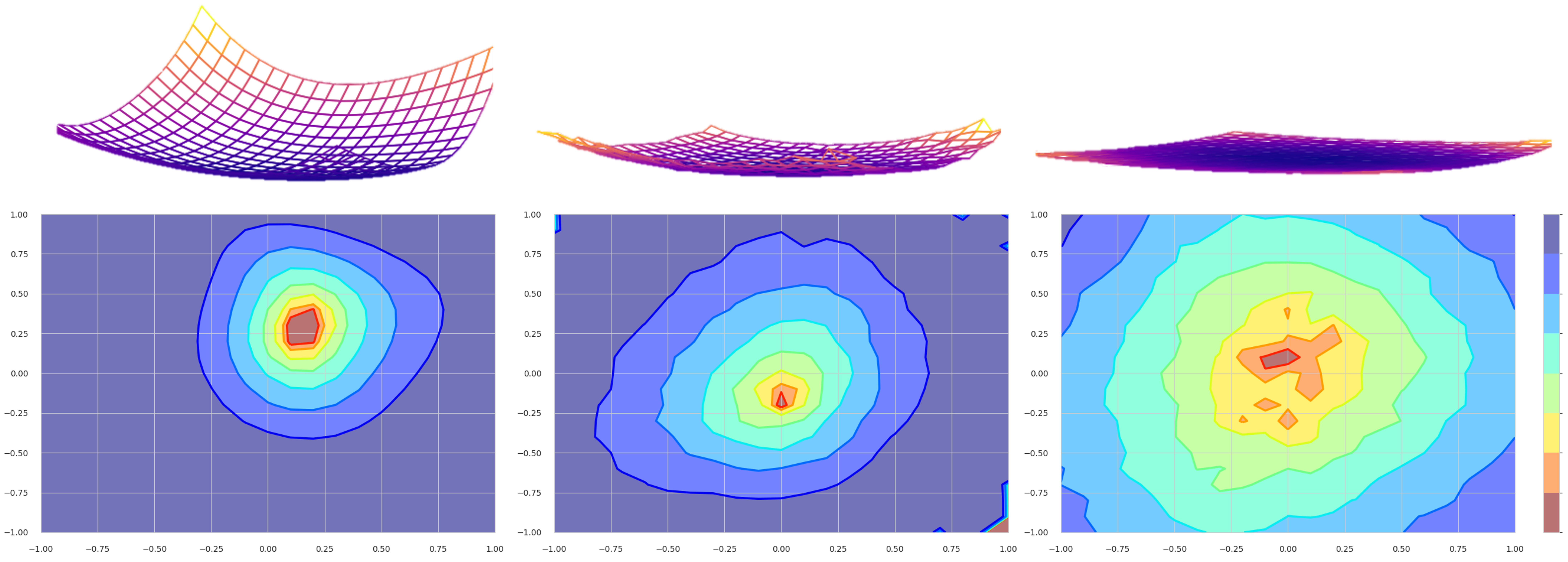}
\caption{The loss surface of a deep network \textit{TwoCNN} on CIFAR10 dataset calculated from l2-regularized cross-entropy loss as a function of network parameters in a two-dimensional subspace. The top rows indicate three-dimensional visualization of loss landscapes and the bottom rows indicates contours of derived loss landscapes \textbf{First column:} loss landscape obtained from \texttt{FedAvg}. \textbf{Second column:} loss landscape obtained from \texttt{SuPerFed-MM}. \textbf{First column:} loss landscape obtained from \texttt{SuPerFed-LM}.}
\label{figureA}
\end{figure*}

\begin{figure*}[!htbp]
\centering
\includegraphics[width=0.8\textwidth]{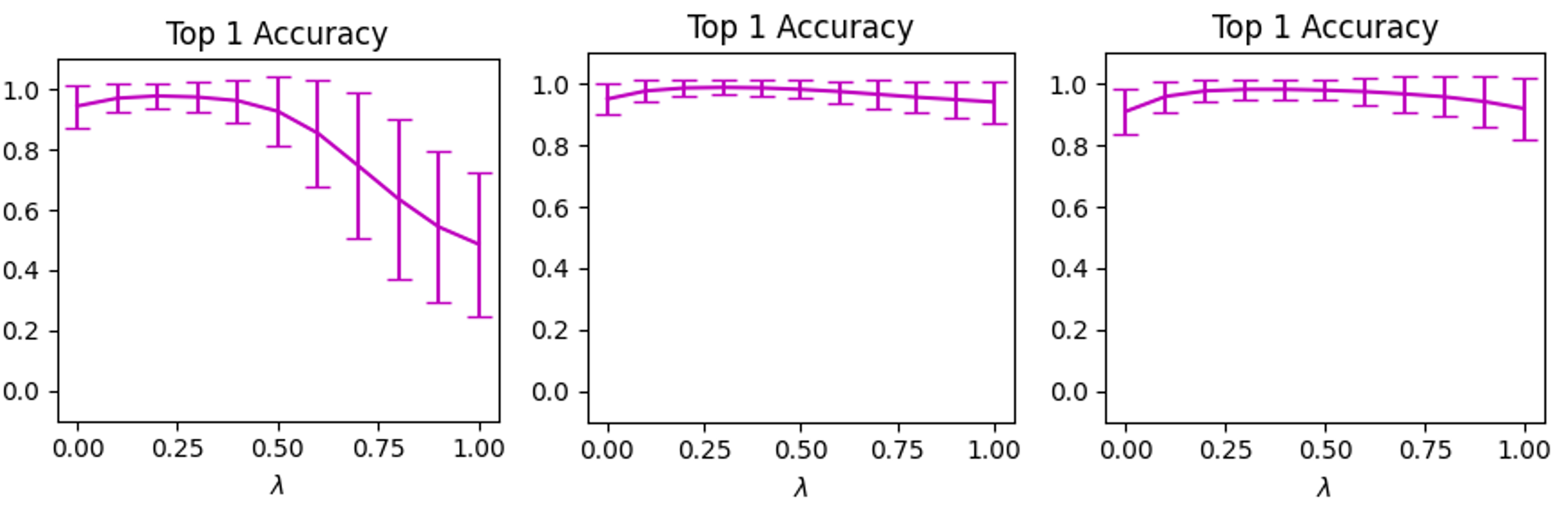}
\caption{Comparison of the personalization performance between \texttt{APFL} (\textbf{left}), \texttt{SuPerFed-MM} (\textbf{middle}), and \texttt{SuPerFed-LM} (\textbf{right}) by varying $\lambda\in[0,1]$. Note that each method is trained on MNIST dataset with \textit{TwoNN} model assuming \textit{pathological non-IID setting} ($\mathrm{K}=500$ and $\mathrm{R}=500$).}
\label{figureB}
\end{figure*}

\section{Appendix}
\subsection{Experimental Details}
Table~\ref{tableA} summarizes the details of each experiment: the number of clients $\mathrm{K}$, total rounds $\mathrm{R}$, local batch size $\mathrm{B}$, number of local epochs $\mathrm{E}$, learning rate $\eta$, and the network. Note that the fraction of sampled clients, $\mathrm{C}$ is adjusted to be 5 clients, and the start of personalization $\mathrm{L}$ is set to 40\% of total rounds. For a stable convergence of a global model, we applied applied learning rate decay by 1\% per each round. Each network architecture is borrowed from the original paper it proposed: \textit{TwoNN}, \textit{TwoCNN}, \textit{StackedLSTM}~\cite{mc+17}, \textit{ResNet}~\cite{resnet}, \textit{MobileNet}~\cite{mobilenet}, and \textit{VGG}~\cite{vgg}. 

\subsection{Loss Landscape of a Global Model} The loss landscape of global models obtained from the proposed methods \texttt{SuPerFed-MM} and \texttt{SuPerFed-LM} are visualized in Figure~\ref{figureA} along with one from \texttt{FedAvg} by following ~\cite{garipov+18}. As the proposed regularization term adjusted by $\nu$ induces connected low-loss subspace, it is natural that the wider minima is acquired from both \texttt{SuPerFed-MM} and \texttt{SuPerFed-LM} compared to that of \texttt{FedAvg}. As the global model learns to be connected with local models in weight spaces per each federation round, flat minima of the global model are induced, i.e., the \textit{connectivity} is observed. 

One thing to note here is that the minima is far wider in layer-wise mixing scheme (i.e., \texttt{SuPerFed-LM}) than that of model-wise mixing scheme (i.e., \texttt{SuPerFed-MM}). It is presumably due to the fact that the diversity of model combinations is richer in the layer-wise mixing scheme than in the model-wise mixing scheme. In other words, the global model should be connected to the larger number of different models. 

Meanwhile, since the global model has wide minima, it can be expected that it is friendly to be exploited by unseen clients even if their data distribution is heterogeneous from existing clients' data distributions. It is because the global model contains many candidates of low-loss solutions thanks to the induced \textit{connectivity}, the chances of finding a suitable model are higher than the global model derived from other FL methods.  

\subsection{Difference from the Existing PFL Methods with Model Interpolation} In previous model mixture-based PFL methods, there exist similar approaches that interpolate the local model and the federated model in the form of a linear combination of a federated model and a local model: (i) \texttt{APFL}~\cite{apfl}, (ii) \texttt{mapper}~\cite{mansour+20} and (iii) \texttt{L2GD}~\cite{l2sgd}.

In (i) \texttt{APFL}, the federated model and the local model are updated \textit{separately} with a fixed $\lambda$, which is not true in \texttt{SuPerFed} as it jointly updates both federated and local models at once. While authors of ~\cite{apfl} proposed the method to dynamically tune $\lambda$ through another step of gradient descent, \texttt{SuPerFed} does not require finding one optimal $\lambda$. It aims to find a good set of models $\mathbf{w}^l$ and $\mathbf{w}^f$ that can be combined to yield good performance regardless of the value of $\lambda$.

In (ii) \texttt{mapper}~\cite{mansour+20}, on the other hand, it iteratively selects only one client (an unrealistic assumption) and requests to find the best local models $\mathbf{w}^l$ and $\lambda$ based on the transmitted global model $\mathbf{w}^g$. The federated model is then updated with the the local model and the optimal $\lambda$ is tuned. In \texttt{SuPerFed}, it is not required to find a fixed value of $\lambda$ and updates of both local and federated models can be completed in a single run of back-propagation using gradient descent. No separate training on each model is required as in other methods. 

In (iii) \texttt{L2GD} proposed by ~\cite{l2sgd}, the interpolation of the global model and the local model is intermittently occurred at the \textit{server}. Moreover, it requires \textit{all} clients to participate in each round, which is a fairly unrealistic assumption in FL.

We only compared with the method applicable for realistic FL settings, \texttt{APFL}. For the comparison, we observed dynamics of performances by varying $\lambda\in[0,1]$ in both \texttt{SuPerFed} (\texttt{SuPerFed-MM} \& \texttt{SuPerFed-LM}) and \texttt{APFL} (See Figure~\ref{figureB}). While \texttt{SuPerFed-MM} and \texttt{SuPerFed-LM} consistently perform well in terms of personalization regardless of the value $\lambda$, \texttt{APFL} performs best only when $\lambda$ is close to 0.25. This supports the evidence of induced \textit{connectivity} between the federated and local models by the regularization. As endpoints (i.e., $\lambda=0$ for a federated model, $\lambda=1$ for a local model) are updated in the way to be connected, both are trained in a mutually beneficial way.

\subsection{Label Noise Scenario}
Since we simulated noisy label scenarios using clean-labeled datasets (MNIST and CIFAR10), we manually obfuscate the labels of training samples following two schemes: pairwise flipping~\cite{pair} and symmetric flipping~\cite{symm}. Both are based on the defining label transition matrix $T$, where each element of $T_{ij}$ indicates the probability of transitioning from an original label $y=i$ to an obfuscated label $\Tilde{y}=j$, i.e., $T_{ij}=p(\Tilde{y}=j|y=i)$. 

\subsubsection{Pairwise Flipping for Label Noise}
The pairwise flipping emulates a situation when labelers make mistakes only within similar classes. In this case, the transition matrix $T$ for a pairwise flip given the noise ratio $\epsilon$ is defined as follows~\cite{pair}.
\begin{align*}
T=\left[\begin{array}{ccccc}
1-\epsilon & \epsilon & 0 & \cdots & 0 \\
0 & 1-\epsilon & \epsilon & \cdots & 0 \\
\vdots & \vdots & \ddots & \ddots & \vdots \\
0 & 0 & \cdots & 1-\epsilon & \epsilon \\
\epsilon & 0 & \cdots & 0 & 1-\epsilon
\end{array}\right]
\end{align*}

\subsubsection{Symmetric Flipping for Label Noise}
The symmetric flipping assumes a probability of mislabeling of a clean label as other labels are uniformly distributed. For $n$ classes, the transition matrix for symmetric flipping is defined as follows~\cite{symm}.
\begin{align*}
T=\left[\begin{array}{ccccc}
1-\epsilon & \frac{\epsilon}{n-1} & \cdots & \frac{\epsilon}{n-1} & \frac{\epsilon}{n-1} \\
\frac{\epsilon}{n-1} & 1-\epsilon & \frac{\epsilon}{n-1} & \cdots & \frac{\epsilon}{n-1} \\
\vdots & \vdots & \ddots & \vdots & \vdots \\
\frac{\epsilon}{n-1} & \cdots & \frac{\epsilon}{n-1} & 1-\epsilon & \frac{\epsilon}{n-1} \\
\frac{\epsilon}{n-1} & \frac{\epsilon}{n-1} & \cdots & \frac{\epsilon}{n-1} & 1-\epsilon
\end{array}\right]
\end{align*} \\

\end{document}